\documentclass[runningheads]{llncs}

\usepackage[T1]{fontenc}

\usepackage{microtype}
\usepackage{graphicx,verbatim}
\usepackage{amsmath}
\usepackage{siunitx}
\usepackage{hyperref}
\usepackage{cleveref}
\usepackage{amssymb}

\usepackage{color}

\usepackage{orcidlink}

\begin{document}

\title{Temporally Consistent Graph Extraction and Matching for Longitudinal Angiographic Images}
\titlerunning{Temporally Consistent Vessel Graph Matching}

\author{Linus Kreitner\inst{1}\orcidlink{0009-0004-5727-6697} \and Laurin Lux\inst{1,2}\orcidlink{0009-0003-7359-6212} \and Carmen Baumann\inst{3}\orcidlink{0000-0001-8662-0582} \and Daniel Rueckert\inst{1,2,4}\orcidlink{0000-0002-5683-5889} \and Martin J. Menten\inst{1,2,4}\orcidlink{0000-0001-8261-7810}} 
\authorrunning{L. Kreitner et al.}
\institute{
Chair for AI in Healthcare and Medicine, Technical University of Munich (TUM) and TUM University Hospital, Munich, Germany \and
Munich Center for Machine Learning (MCML), Munich, Germany \and
Ophthalmology, Technical University of Munich, Munich, Germany \and
Department of Computing, Imperial College London, UK \\
\email{\{linus.kreitner, martin.menten\}@tum.de}}
  
\maketitle 

\begin{abstract}
Recent advances in angiographic imaging have enabled longitudinal visualization of the microvasculature. Image processing pipelines based on vessel graphs are able to resolve subtle temporal changes at the level of individual blood vessels. However, current strategies for graph extraction, refinement, and matching are highly sensitive, with even minuscule differences in the underlying segmentation map resulting in substantially different vessel graphs. These artifacts severely inhibit the ability to accurately match sequential vessel graphs of the same subject over time. To address this problem, we propose a strategy that matches graphs before jointly refining them. Specifically, we perform an early matching after basic graph extraction before removing spurious bulges and merging junctions in both graphs using joint information. In experiments with complex retinal vessel graphs, we demonstrate that this strategy results in a higher matched area without graph fragmentation compared to separate or no refinement, respectively.


\end{abstract}

\section{Introduction}
\label{sec:intro}

Advances in medical imaging technology have made detailed visualization of the microvasculature a clinical reality \cite{hartung2011magnetic,oglat2018review,spaide2018optical}. The advent of non-invasive, cost-effective angiographic imaging has resulted in the emergence of longitudinal datasets that monitor growth, regression, and remodeling of the vasculature \cite{ai2024ai,kim2026longitudinal,riedel2024isles}. Due to the complexity of the underlying anatomy, clinicians and researchers rely on automated image processing algorithms to extract quantitative vascular biomarkers. As a first step, these tools usually delineate the blood vessels \cite{moccia2018blood,ikram2013retinal,li2024ocular,arsalan2022detecting}. The resulting segmentation map already facilitates calculation of basic biomarkers, such as vessel density, and their changes over time.

However, with improving imaging capabilities there is a growing interest in more granular biomarkers that resolve temporal changes at the level of individual blood vessels. Such fine-grained analysis requires advanced image processing pipelines that extract and process vessel graphs from angiographic images \cite{kirbas2003vessel}. Vessel graphs are typically constructed by obtaining a topological representation of a segmentation mask and converting it to a graph that compactly encodes bifurcation points and connecting vessel segments as nodes and edges \cite{li20213d,muller2024survey,yu2022vessel,fhima2022pvbm}. Additionally, nodes and edges can be imbued with additional features, such as vessel thickness, segment length, curvature, or branching angle \cite{drees2021scalable,paetzold2021whole}. Afterwards, this basic graph can be iteratively refined by removing superfluous nodes and edges. In the longitudinal setting, graph matching is required to establish correspondences between edges that represent the same anatomical structures at different time points \cite{feuillatre2015improved,fang2019greedy}.
However, current strategies for graph extraction, refinement, and matching are highly sensitive to small variations in the input segmentation mask. Even minuscule differences can result in substantially different vessel graphs. These artifacts severely inhibit the ability to accurately match sequential graphs of the same subject, even if a clear correspondence between the vasculature at two time points exists. Currently, these matching errors greatly diminish the utility of vessel graphs for analysis of longitudinal angiographic imaging data.

\begin{figure}[htbp]
    \centering
    \includegraphics[width=0.80\linewidth]{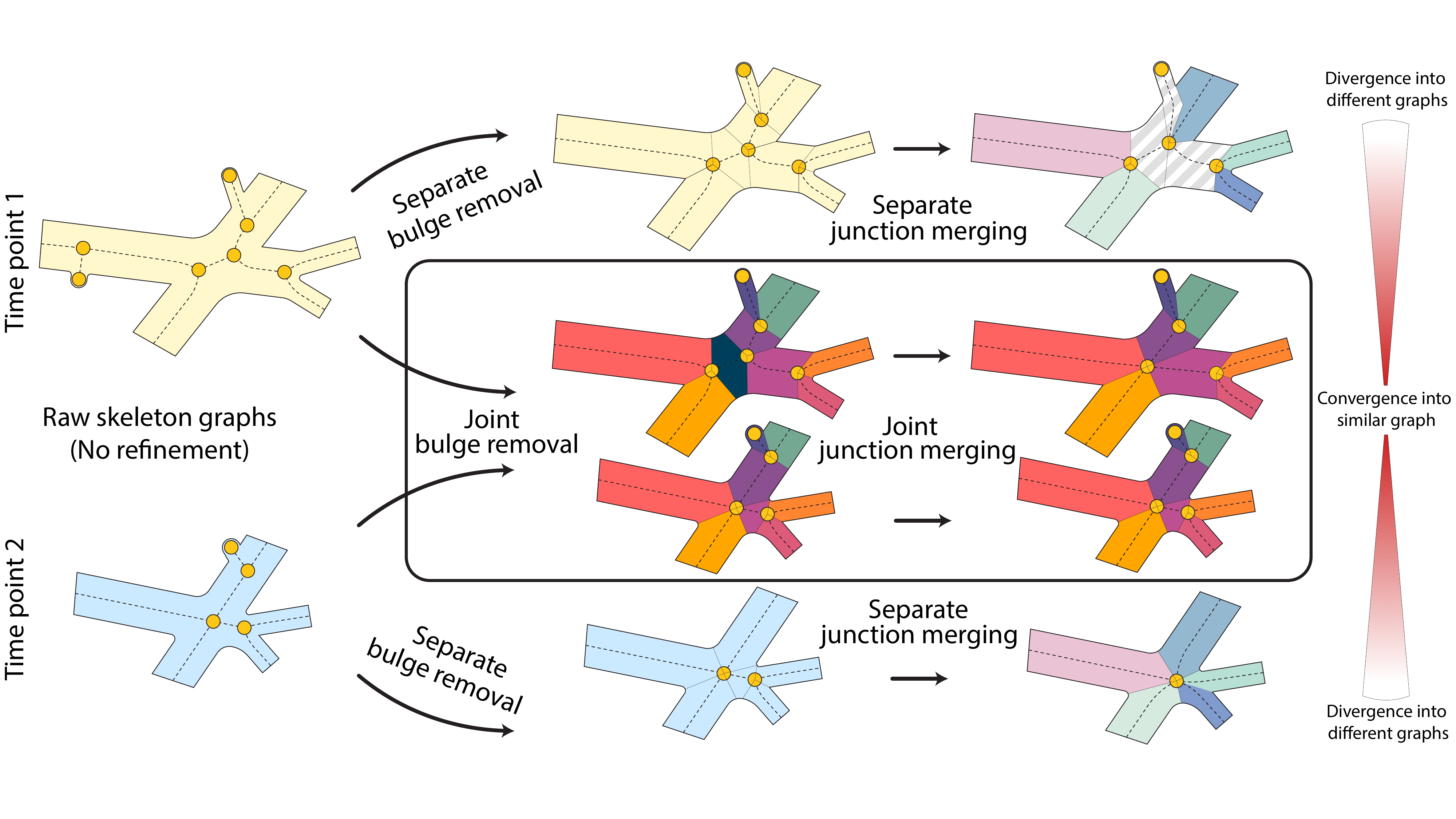}
    \caption{
    Pipeline to refine and match two vessel graphs from 2D segmentation masks of longitudinal angiographic images.
    We propose early matching of temporally adjacent graphs and refinement using joint information to promote convergence into a similar graph representation.
    }
    \label{fig:graph_extraction}
\end{figure}

Our work addresses this problem by making a series of contributions:
\begin{enumerate}
\item We show that highly similar angiographic images can yield vastly different vessel graphs. We establish that small variations in the underlying segmentation map have an outsized impact on the graph refinement stage.
\item To address this problem, we propose to refine longitudinal vessel graphs under consideration of temporally adjacent samples. Specifically, we perform an early matching after basic graph extraction before removing spurious bulges and merging junctions in both graphs using joint information (see \Cref{fig:graph_extraction}).
\item In experiments using a unique optical coherence tomography angiography (OCTA) dataset with repeated intra-subject scans, we demonstrate the utility of our method in a highly challenging setting that requires matching longitudinal vessel graphs with 2,000 edges.
\end{enumerate}

\section{Methods}

We propose an early graph matching and joint refinement strategy for better graph correspondence between time points. We first introduce the simpler "separate refinement" graph processing pipeline that sequentially extracts (\Cref{sec:graph_extraction}), refines (\Cref{sec:refinement}), and matches (\Cref{sec:matching}) two time-adjacent graphs. The individual steps build on well-established principles such as skeleton extraction, topological simplification, and artifact removal, and we show that the pipeline performs comparably to widely adopted baseline methods\cite{voreen,BUMGARNER2022100189,Moriconi2018Elastic}. We then show how to adapt this baseline for our proposed strategy of early graph matching and joint refinement (\Cref{sec:joint}). Controlling the entire pipeline allows us to quantify the benefit of our joint refinement strategy without extraneous factors affecting the results.

\subsection{Graph extraction}
\label{sec:graph_extraction}

We define a vessel graph $G=(V,E)$ as a set of nodes $V$ denoting bifurcation points as well as a set of edges $E$ representing the connecting vessels.

\noindent\textbf{Node extraction} To construct the graph, we initially extract the skeleton of the 2D segmentation mask \cite{LEE1994462}. We define all centerline pixels with a single 8-connected neighbor as leaf nodes and pixels with more than two neighbors as junction nodes. All adjacent junction nodes are grouped into hubs before applying a breadth-first search to find all distinct paths between hubs. 

\noindent\textbf{Edge assignment} We extract the distinct paths (edges) between hubs as the shortest 8-connected pixel walks between two hubs that do not visit another hub in between. All junction nodes within a hub are merged into a single merged node and placed at the centroid of the hub. We then remove redundant skeleton pixels around the merged node that no longer lie on the shortest path, or add pixels to reconnect the incoming edges if the centroid is not on the skeleton.

\noindent\textbf{Node and edge features} Finally, we compute the radius $r$ of both nodes and edges as the exact distance transform (EDT) at their coordinates or the median EDT at their skeleton coordinates, respectively. Additionally, each edge carries its underlying skeleton and surrounding 2D segmentation mask pixels as features.

\subsection{Graph refinement}
\label{sec:refinement}

The previously described graph extraction results in a raw graph that still contains many small spurious edges due to small variations in the segmentation mask. To separate true vessels from these artifacts, we conduct three additional refinement steps to simplify the vessel graph topology.

\noindent\textbf{Bulge removal}
This first refinement step aims to remove spurious edges $e$ leading from a base node $n_b$ towards a leaf node $n_l$. We use the resolution-agnostic bulge size metric defined by Drees et al. \cite{drees2021scalable},
\begin{equation}
    \beta(e)= \frac{\text{length}(e)-\text{radius}(n_b)+\text{radius}(n_l)}{\text{radius}(e)},
\end{equation}
to remove any bulges with a size smaller than a threshold $T_\text{bulge}$.

\noindent\textbf{Adjacent node merging}
When two vessels cross approximately at a 90° angle, a degree-4 crossing node is created. However, if two thick vessels overlap at an acute angle, the medial axis often produces an H-shaped configuration consisting of two degree-3 junction nodes connected by a short bridging edge $e_H$. The second refinement step resolves this artifact by merging these two junction nodes into a single higher-degree crossing node. The cost of this merging operation consists of two terms. The first penalizes long, thin bridges and is defined as $\mathcal{C}_\text{merge}^{e_H}=\frac{\text{length}(e_H)}{\text{radius}(e_H)}$. A second term
measures geometric plausibility by quantifying how well incoming edges can be paired with outgoing edges of similar orientation and radius. 
Crossings involving more than two junction nodes are resolved by iteratively merging node pairs starting with the lowest merge cost.

\noindent\textbf{Skeleton cleanup}
Merging junction nodes and placing a new crossing node at the center of their connecting edge leads to locally suboptimal skeleton geometry. In particular, the incident branches may share pixel segments near crossings. This final refinement step rewires the skeleton paths around crossing nodes to minimize the overlap of their incoming edges. We remove all skeleton pixels within the node's radius and then reconnect incoming edges by a straight line.

\subsection{Graph matching}
\label{sec:matching}

\begin{figure}[htbp]
    \centering
    \includegraphics[width=0.80\linewidth]{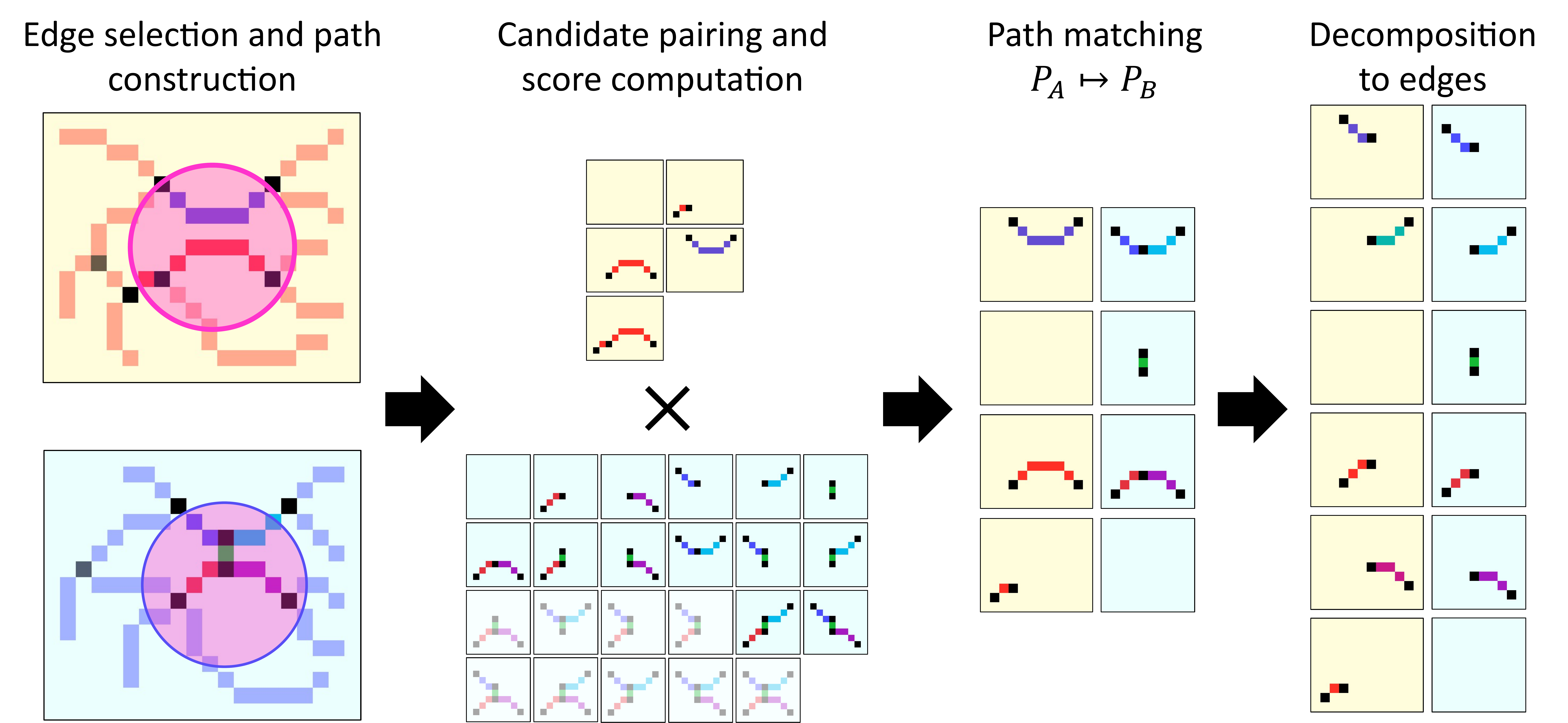}
    \caption{Four-step graph matching process to establish a 1-to-1 pairing between edges.}
    \label{fig:matching}
\end{figure}

Graph matching aims to find a correspondence between the edges of two related graphs by establishing a bijective mapping $\mathcal M: E_A \mapsto E_B$ that explicitly assigns each edge in graph A to exactly one edge in graph B \cite{kong1989digital}. However, such a mapping is unlikely to exist for vessel graphs. For instance, intermediate bulges in one graph can splinter edges into multiple smaller edges, preventing a 1-to-1 correspondence. We therefore relax the strict matching objective in favor of a more flexible path mapping $\mathcal M': P_A \mapsto P_B$, with $P=\{(e_1,\dots,e_n): e\in E\}$ and $n\in \mathbb N_0$ (see \Cref{fig:matching}). A path is a concatenation of multiple edges with a single start and single endpoint, where each edge may only appear once. The arc length of a path is defined as the cumulative Euclidean distance between neighboring skeleton pixels.

\noindent\textbf{Edge selection and path construction}
To find the optimal mapping between two graphs, we initially define a sparse cost matrix between paths in graph A and graph B. To avoid computing the match affinity for all possible $2^{|E_A|}\times2^{|E_B|}$ path combinations, we only consider the most likely candidates.
One of the most important factors for similarity is spatial distance. We therefore align both graphs by computing the global affine transformation that maps each pixel coordinate in graph B to the respective coordinate in graph A.

To build the candidate set for a given edge $e_A$ in graph A, we find all edges in graph B that are within a given search distance of $e_A$. 
We generate all possible paths by concatenating edges of the selected subset. Depth-first path exploration stops if we encounter unrealistic angles or radius changes between edges. We then analogously generate all candidate paths in graph A by finding all possible paths that include $e_A$ and consist of edges that are within search distance.

\begin{figure}[t]
    \centering
    \includegraphics[width=0.95\linewidth]{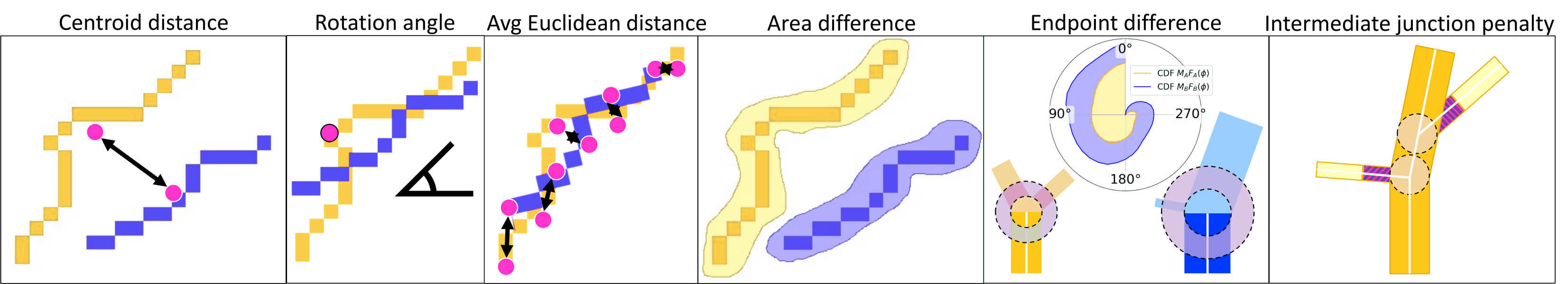}
    \caption{The six metrics used to calculate the match cost between candidate paths.}
    \label{fig:graph_matching}
\end{figure}

\noindent\textbf{Pairing and score computation}
We compute a match cost $s_c$ for each candidate pairing $c=(p_A, p_B)$ as the weighted sum of the following metrics (see \Cref{fig:graph_matching}):
\begin{enumerate}
    \item Distance: Euclidean distance between the path centroids.
    \item Rotation: Acute rotation angle $|\phi|\in[0,\ang{90}]$ to align path B with path A.
    \item Shape: Pairwise Euclidean distance between the $K$ points of the resampled paths that were aligned using the found centroids and rotation angles.
    \item Area: Defined by $a_p=\text{arclength}_p \times r_p$.
    \item Endpoint: We compare the start and endpoints of two paths by computing the earth moving distance between the nodes' angular densities $\rho(n,\phi)$. The density is given by counting non-zero mask pixels in a ring around the node with outer radius $r_\text{out}=2\cdot r_\text{in}=2\cdot r_n$ and dividing by the total pixel count.
    \item Intermediate junction penalty: The sum of penalties for each node within a path, where each protruding edge adds to a node's respective penalty proportional to its radius and length.
\end{enumerate}

\noindent\textbf{Path matching and decomposition}
After computing the match cost for each candidate pair $c\in C$, we define the following Boolean decision variables: $x_c$ signals whether a candidate is selected, while $m_e^G = \sum_{c \in C: e\in E^G_c} x_c$ with $G\in\{A,B\}$ signals whether a given edge in graph A or B is matched. Here $E^G_c$ denotes the set of edges that are part of the candidate $c$ in graph $G$. We then ensure at most one candidate for a given edge may be selected by enforcing $\sum_{c\in C:e\in E_c} x_c \leq 1$. Finally, we define the penalty for an edge $e$ remaining unmatched by a linear function $\lambda a_e+\epsilon$ of its area $a_e$. The matching objective is then given by
\begin{equation}
    \min \underbrace{\sum_{c\in C} s_cx_c}_\text{match cost} + \underbrace{\lambda \left(\sum_{e\in E_A} a_e(1-m^A_e) + \sum_{e\in E_B} a_e(1-m^B_e) \right) + \epsilon}_\text{unmatched penalty}.
\end{equation}
We solve this binary integer linear program using PuLP's time-constrained COIN-OR Branch and Cut (CBC) solver \cite{forrest2024coin}.

\noindent\textbf{Decomposition into edges}
Finally, the resulting path matches must be decomposed into 1-to-1 matches of individual edges. For this, we walk along both paths simultaneously and whenever we encounter a junction in one path, we insert a degree-2 node in the other path. These degree-2 nodes manually fragment edges to achieve direct one-to-one correspondence. 

\subsection{Joint refinement of matched graphs}
\label{sec:joint}

When performing graph refinement independently on adjacent longitudinal graphs, small structures may be removed in one graph because they fall below the bulge removal threshold, even though the structure is clearly present at other time points. To improve matching correspondence across time points, we therefore propose to perform the graph refinement steps described in \Cref{sec:refinement} using information from other graphs.

Specifically, we first perform 1) basic graph extraction separately, then 2) conduct the matching phase on these raw graphs, and finally perform  3.1) bulge removal and 3.2) junction merging using joint information.
A bulge edge in graph A is removed if and only if both its bulge size and that of its matched counterpart in graph B are below the bulge removal threshold. Unmatched edges are treated as before, without additional constraints. Similarly, we only collapse an H-bridge edge in one graph if the matched edge in the other graph also satisfies the removal criterion. By coupling refinement decisions across time points, this strategy suppresses artifacts caused by imperfect segmentation errors while preserving anatomically consistent vessel structures.

\section{Experiments and Results}

\begin{figure}[htbp]
    \centering
    \includegraphics[width=0.95\linewidth]{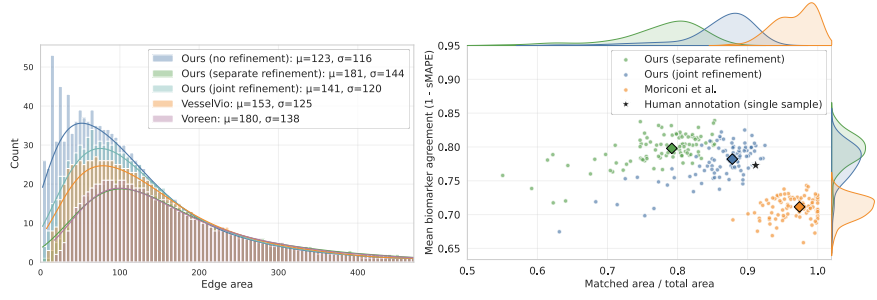}
    \caption{
    Quantitative comparison of the introduced graph construction methods. Left: Graph extraction without refinement results in graphs with spurious edges that do not reflect the true underlying physiology. Graph refinement in our pipeline and established baselines reduces the number of spurious edges.
    Right: Joint refinement results in a higher matched area compared to separate refinement, while exhibiting a similar or better edge matching quality than separate refinement or the Moriconi et al. baseline.
    }
    \label{fig:quantitative_results}
\end{figure}

\noindent\textbf{Longitudinal OCTA dataset} For our experiments, we use an in-house dataset of 160 retinal OCTA images. Two repeated scans of 80 healthy eyes were acquired on the same day. For each scan, we extract 1.5 $\times$ 1.5 mm$^2$ large en-face projections of the inner retinal vasculature. We obtain a detailed $608 \times 608$ pixel segmentation mask of all images using a public tool for OCTA image segmentation \cite{kreitner2024synthetic}. As the vascular structure of the retina is highly stable over this short time interval, we expect only minor structural changes due to varying light exposure and the circadian cycle. This idealized setting allows us to isolate the effect of small artifacts inherent to the graph construction pipeline.

\begin{figure}[t]
    \centering
    \includegraphics[width=0.775\linewidth]{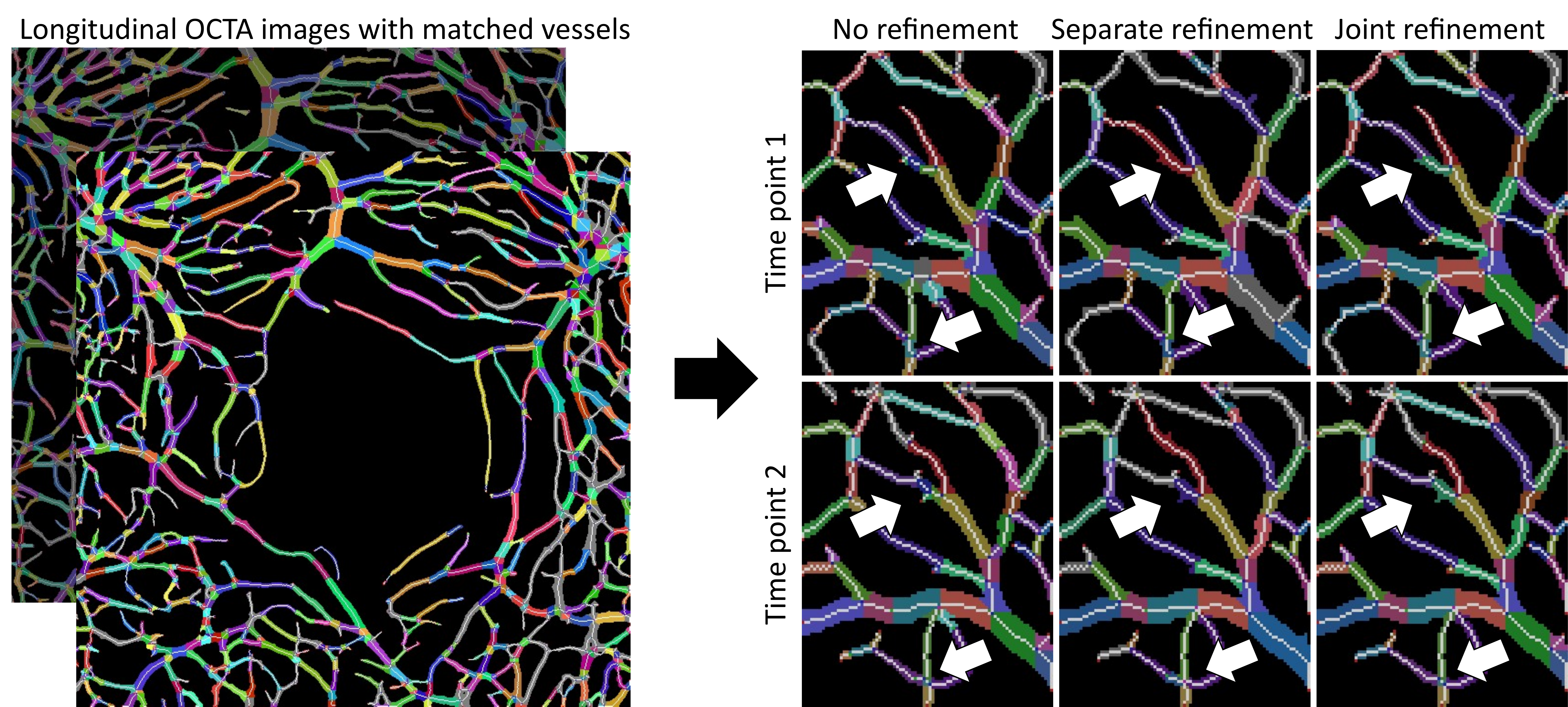}
    \caption{Representative example of matching quality for the three different graph refinement strategies. No refinement fragments edges due to spurious branches (upper arrow), while refining graphs separately causes poor matching quality, indicated by an increased number of unmatched or wrongly paired edges (lower arrow). In contrast, joint refinement yields convincing edge-level matches.}
    \label{fig:qualitative_results}
\end{figure}

\noindent\textbf{Graph refinement to reduce spurious edges} We compare the three introduced graph construction strategies: i) graph extraction without refinement, ii) extraction with separate refinement, and iii) extraction with joint refinement. Additionally, we include two established graph extraction tools, Voreen \cite{voreen} and VesselVio \cite{BUMGARNER2022100189} for comparison. First, we demonstrate the benefit of graph refinement when processing complex vessel graphs. Small segmentation variations along vessel boundaries can result in spurious bulges. Unrefined graphs, on average, contain about 1,300 edges, the majority of which are very small segments (see \Cref{fig:quantitative_results,fig:qualitative_results}). Our bulge removal and junction merging refinement reduces this fragmentation and yields a more coherent graph topology. We find that our graph extraction strategy yields similar graphs as established baseline methods with regard to number of edges and average vessel size. 

\noindent\textbf{Benefit of our joint graph refinement} However, refining longitudinal scans separately can cause the resulting graphs to diverge topologically, ultimately degrading matching performance. To quantify this effect, we compare matched area and matching quality on separately refined graphs using both our method and the Moriconi et al. baseline \cite{Moriconi2018Elastic}, and contrast these results with our proposed joint refinement approach. We define matching quality $Q=1-\operatorname{sMAPE}(i,j)$, with $\operatorname{sMAPE}(i,j)
=\frac{1}{|\mathcal{M}|}\sum_{(i,j)\in\mathcal{M}}\frac{|b_i-b_j|}{|b_i|+|b_j|}$ \cite{FLORES198693}, as the mean agreement of vessel-specific biomarkers $b$ (arclength, radius, area, and straightness) of all matched vessels $i$ and $j$. Thus, biomarker agreement guards against trivially increasing coverage though random pairs. We tune all methods on a manually annotated graph pair.

Separate refinement decreases the total area of the graph that can be successfully matched (see \Cref{fig:quantitative_results} and \Cref{fig:qualitative_results}). In contrast, performing early matching followed by joint refinement increases the total matched area, while maintaining stable match quality. The baseline produces a larger number of matches, however, the biomarker correspondence between matched vessels is substantially lower.

\section{Discussion}

Longitudinal angiographic imaging combined with vessel graphs has the potential to detect temporal changes at the level of individual blood vessels.
However, this study has shown for the first time that current strategies for graph extraction and refinement are highly sensitive to small variations in input images. These artifacts severely inhibit the ability to successfully match sequential vessel graphs of the same subject, even if a clear correspondence between the vasculature exists. To address this problem, we have introduced a strategy that matches graphs before joint refinement. In experiments with complex retinal vessel graphs, we have demonstrated that this strategy results in a higher matched area without graph fragmentation compared to separate or no refinement, respectively.

These results were obtained in a carefully crafted setting with a specifically developed graph extraction and matching algorithm. Nonetheless, we postulate that our strategy of early matching before refinement is conceptually compatible with other methods, such as those that construct graphs by predicting links between a candidate set of nodes \cite{shin2019deep,yu2022vessel} or directly extract the graph from the image \cite{berger2025cross,shit2022relationformer}. Similarly, our method can be applied to three-dimensional images or to the simultaneous extraction of more than two vessel graphs. As such, our work has the potential to support the extraction of advanced graph-based biomarkers from longitudinal angiographic images.

\newpage
\begin{credits}
\subsubsection{\ackname}
This work was partially funded by ERC Grant Deep4MI (Grant No. 884622). Martin J. Menten is
funded by the German Research Foundation under projects 528297171 and 532139938.

\subsubsection{\discintname}
The authors have no competing interests to declare that are relevant to the content of this article.
\end{credits}

\bibliographystyle{splncs04}
\bibliography{bibliography}

\end{document}